\documentclass[10pt,twocolumn,letterpaper]{article}

\usepackage[applications]{wacv}

\usepackage{float}
\usepackage{capt-of}
\usepackage[ruled,lined]{algorithm2e}
\makeatletter
\@ifundefined{linenomath}{\newenvironment{linenomath}{}{}}{}
\makeatother

\DontPrintSemicolon
\SetAlgoLongEnd
\SetKwInOut{KwIn}{Input}
\SetKwInOut{KwOut}{Output}
\SetKwInOut{KwInit}{Initialize}
\SetKwComment{tcp}{// }{}
\SetAlgoCaptionSeparator{ }
\SetAlCapSkip{0.35em}
\SetInd{0.45em}{0.8em}

\DontPrintSemicolon
\SetAlgoLongEnd
\SetKwInOut{KwIn}{Input}
\SetKwInOut{KwOut}{Output}
\SetKwInOut{KwInit}{Initialize}
\SetAlgoCaptionSeparator{ }
\SetAlCapSkip{0.35em}
\SetInd{0.45em}{0.85em}

\definecolor{wacvblue}{rgb}{0.21,0.49,0.74}
\usepackage[breaklinks,colorlinks,allcolors=wacvblue]{hyperref}

\def\wacvPaperID{*****}
\def\confName{WACV}
\def\confYear{2027}

\title{Shell-Supervised Gaussian Splatting for Urban Real-to-Sim Reconstruction}

\author{
\begin{tabular}{c}
Yuan Yang$^{1}$ \quad
Peijun Lu$^{2}$ \quad
Fangzhou Lu$^{1}$ \quad
Sai Fan$^{1}$ \quad
Siqi Yan$^{1}$ \quad
Chenyuan Zhang$^{1}$ \\
Haobo Liang$^{1,\dagger}$ \quad
Yicheng Wang$^{1,\dagger}$ \\
{\small $^{1}$Hong Kong Center for Construction Robotics \quad
$^{2}$Tsinghua University} \\
{\tt\small yuanyang@ust.hk \quad hbliang@ust.hk$^{\dagger}$ \quad yichengwang@ust.hk$^{\dagger}$}
\end{tabular}
}

\begin{document}
\maketitle

\begin{abstract}
Real-to-sim reconstruction for embodied AI requires geometry that is useful for collision reasoning, navigation, and agent-environment interaction, not only photorealistic novel-view synthesis. However, close-range urban facades are difficult for video-to-3D reconstruction: glass, reflections, repeated windows, and weak texture can produce visually plausible renderings with unstable surface geometry. We introduce shell-supervised Gaussian Splatting, a reconstruction-stage framework that uses an external facade structural shell as lightweight geometric supervision for video-driven Gaussian reconstruction. The method aligns an exterior shell to the video reconstruction frame, renders per-view depth, camera-space normal, and valid-mask maps, and applies these cues through mask-gated losses during Gaussian optimization. This design preserves RGB-driven appearance while regularizing only visible shell-supported facade regions. Experiments on anonymized close-range urban facade scenes show improved facade orientation and visible-surface point-cloud consistency over photo-only, monocular-cue, and surface-oriented Gaussian baselines, while maintaining comparable held-out rendering quality.

\end{abstract}

\section{Introduction}

Real-to-sim reconstruction is becoming an important direction for
embodied AI: instead of manually designing synthetic environments,
recent systems convert real-world videos into interactive simulation
assets for agent training~\cite{Xie2025Vid2Sim}. In this setting, a
reconstruction is not only a novel-view synthesis model, but also the
geometric substrate for collision reasoning, physical interaction, and
navigation-relevant observations.

Recent neural rendering methods, especially 3D Gaussian Splatting
(3DGS), provide real-time photorealistic reconstruction from posed
images~\cite{Kerbl2023GaussianSplatting}. However, photorealistic
rendering does not guarantee reliable geometry. This gap is especially
problematic for close-range urban facades, where glass, reflections,
repeated windows, weak texture, and limited viewpoints can yield
plausible RGB renderings but inconsistent surface orientation.
Reflective and transparent surfaces are also known to be difficult for
3D reconstruction systems~\cite{Whelan2018MirrorGlass}. For urban real-to-sim pipelines, these geometric errors can affect the physical and navigational usefulness of the reconstructed scene.

Existing geometry-aware reconstruction methods address this problem
from different angles. Depth and normal priors can regularize Gaussian
Splatting~\cite{Turkulainen2025DNSplatter}, surface-oriented Gaussian
representations such as 2DGS improve geometric consistency by changing
the primitive representation~\cite{Huang2024TwoDGS}, and
building-prior methods show that external building geometry can guide
Gaussian reconstruction~\cite{Zhang2025GS4Buildings}. Nevertheless,
close-range facades remain challenging because monocular cues are still
inferred from ambiguous RGB evidence, while detailed building priors can be too sensitive or difficult
to release.

This work introduces a facade shell-guided Gaussian reconstruction
framework for the video-to-3D stage of urban real-to-sim pipelines. The
central observation is that a complete semantic building model is
unnecessary: the method only requires a lightweight exterior structural
shell that captures visible facade geometry. Such exterior geometry can
be obtained from existing exterior models, laser scanning, street-level
mapping, or photogrammetric reconstruction~\cite{Rashdi2022ScanToBIM}.
After aligning the shell to the video reconstruction frame, we render
per-view shell depth, camera-space normals, and valid masks for
mask-gated geometric supervision during 3DGS optimization. The shell acts
as supervision rather than initialization or replacement of the
video-derived reconstruction, allowing the method to preserve
video-driven appearance while regularizing only shell-supported facade
regions.

Experiments are conducted on two anonymized close-range urban facade
sequences in a Vid2Sim-style reconstruction setting. The primary scene
provides the full comparison against photo-only 3DGS, monocular-cue
supervision, and 2DGS baselines, while the secondary scene serves as a
cross-scene validation case. Across these evaluations, shell-guided
supervision improves facade-orientation consistency and visible-surface
point-cloud accuracy on shell-supported regions while preserving
comparable held-out rendering quality. These results suggest that
lightweight exterior facade shells provide useful geometric supervision
for video-driven urban facade reconstruction without requiring a complete
semantic building model.

Our contributions are summarized as follows:
\begin{itemize}
\item We formulate shell-supported facade geometry consistency as a
reconstruction-stage objective for urban real-to-sim pipelines, where
photorealistic rendering alone is insufficient.
\item We introduce an exterior facade shell as lightweight geometric
supervision for Gaussian reconstruction, avoiding the need for a complete
semantic building model.
\item We render the aligned shell into per-view depth, camera-space
normal, and valid-mask supervision, and show improved visible facade
geometry over photo-only, monocular-cue, and surface-oriented Gaussian
baselines.
\end{itemize}

\section{Related Work}

\paragraph{Neural rendering and Gaussian scene reconstruction.}
Neural rendering enables photorealistic view synthesis from posed
images, from NeRF-style volumetric radiance fields
~\cite{Mildenhall2020NeRF} to scalable and efficient variants for
unbounded or large-scale scenes
~\cite{Barron2022MipNeRF360,Muller2022InstantNGP,Tancik2022BlockNeRF,Turki2022MegaNeRF,Rematas2022UrbanRF}.
Street-view neural rendering further studies urban captures with sparse
view overlap, large-scale backgrounds, and dynamic driving scenes
~\cite{Xie2023SNeRF,Guo2023StreetSurf}. 3D Gaussian Splatting provides a real-time explicit
representation for photorealistic rendering~\cite{Kerbl2023GaussianSplatting}.
Subsequent Gaussian methods improve robustness and
structure through anti-aliasing, anchor-based organization,
density control, surface-aligned representations, opacity
fields, surfels, planar primitives, or street-scene
formulations~\cite{Yu2024MipSplatting,Lu2024ScaffoldGS,Zhang2024PixelGS,Guedon2024SuGaR,Huang2024TwoDGS,Yu2024GOF,Jiang2025GaussianSurfels,Zanjani2025PlanarGS,Yan2024StreetGaussians}.
Our work keeps a video-driven 3DGS backbone, but adds
exterior facade-shell supervision for close-range urban
facade geometry.

\paragraph{Geometric priors for neural reconstruction.}
Photometric supervision alone often leaves geometry under-constrained.
Prior work improves reconstruction through learned video geometry
~\cite{Sun2021NeuralRecon}, implicit surface formulations
~\cite{Yariv2021VolSDF,Wang2021NeuS,Li2023Neuralangelo}, sparse depth
or sparse-view regularization~\cite{Deng2022DSNeRF,Niemeyer2022RegNeRF},
monocular depth/normal cues~\cite{Yu2022MonoSDF}, and structural or
normal priors for weakly textured and planar regions
~\cite{Guo2022ManhattanSDF,Wang2022NeuRIS}. DN-Splatter shows that
depth and normal priors can regularize Gaussian Splatting geometry and
improve meshing~\cite{Turkulainen2025DNSplatter}. In contrast to
image-derived or sensor-derived priors, our supervision is rendered
from an aligned exterior facade shell.

\paragraph{Real-to-sim and embodied simulation.}
Embodied AI research relies on simulation platforms for scalable
training and evaluation, including Matterport3D, Gibson, Habitat,
AI2-THOR, iGibson, and CARLA
~\cite{Chang2017Matterport3D,Xia2018GibsonEnv,Savva2019Habitat,Kolve2017AI2THOR,Shen2021iGibson,Dosovitskiy2017CARLA}.
Recent real-to-sim systems construct interactive environments directly
from real-world captures, such as Video2Game and Vid2Sim
~\cite{Xia2024Video2Game,Xie2025Vid2Sim}. Our work targets the
reconstruction stage of such pipelines: rather than evaluating
downstream reinforcement learning, we focus on asset-level geometry
needed before collision-aware simulation can be reliable.

\paragraph{Building priors, facade shells, and reflective surfaces.}
External structural information can improve reconstruction
and localization in built environments. Prior work on
scan-based building modeling and mapping shows that
exterior geometry can be obtained from laser scanning,
mobile mapping, aerial mapping, or photogrammetry~\cite{Rashdi2022ScanToBIM}.
Architectural-prior methods further use structural building
geometry for LiDAR-camera pose refinement~\cite{Vega2024BIMCaP}, and
GS4Buildings guides Gaussian reconstruction with semantic 3D building
models~\cite{Zhang2025GS4Buildings}. Unlike GS4Buildings, which uses
semantic building models for building-level Gaussian reconstruction, our
method keeps a video-driven SfM/Vid2Sim initialization and uses a
lightweight exterior shell only as mask-gated per-view supervision for
close-range handheld facade capture.

\begin{figure*}[t]
\centering
\includegraphics[width=\textwidth,trim=0 6 0 6,clip]{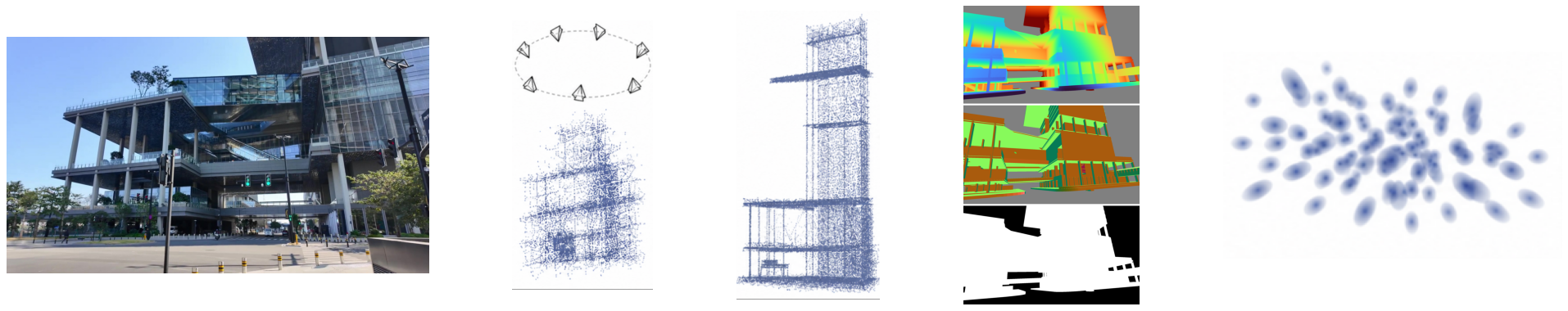}
\caption{Overview of the proposed shell-supervised Gaussian reconstruction framework. A monocular facade video is first reconstructed with an SfM / Vid2Sim-style Gaussian initialization. An aligned exterior facade shell is then rendered into per-view depth, camera-space normal, and valid-mask supervision. These shell-rendered cues are applied only on valid shell-supported pixels during Gaussian optimization, preserving video-driven appearance while improving visible facade geometry.}
\label{fig:pipeline}
\end{figure*}

\section{Method}

Given a handheld monocular video of an urban facade, our
goal is to optimize a video-driven 3D Gaussian scene while
improving the geometry of shell-supported facade regions. The input consists of RGB frames $I_1,\ldots,I_N$, camera
intrinsics and poses $(K_i,R_i,t_i)$ for each frame, estimated by an
SfM / Vid2Sim-style video-to-3D
pipeline~\cite{Schoenberger2016SFM,Xie2025Vid2Sim}, and an external
facade structural shell $S$. The shell is an exterior-only
geometric prior and is converted into per-view supervision rather than
used as a complete semantic building model.

Our method has three stages, as shown in Fig.~\ref{fig:pipeline}. First,
we prepare and validate a fixed shell-to-camera alignment that places the
facade shell in the same coordinate frame as the video reconstruction.
Second, we render the aligned shell into each training view to obtain
shell depth, shell normal, and valid-mask supervision. Third, we optimize
a 3D Gaussian representation~\cite{Kerbl2023GaussianSplatting} with RGB
photometric supervision and mask-gated shell losses. The shell losses are
applied only on valid shell-supported pixels, avoiding constraints in
sky, dynamic, occluded, or out-of-shell regions. The alignment is treated as a validated geometric prior rather than an
absolute metric reference, and is kept fixed during optimization.

\subsection{Vid2Sim-style Gaussian Reconstruction}

We use a 3DGS-style reconstruction backbone within the Vid2Sim-style
video-to-3D setting~\cite{Xie2025Vid2Sim}. Neural rendering methods such
as NeRF demonstrate that posed images can be optimized into
photorealistic view-synthesis models~\cite{Mildenhall2020NeRF}. We
follow the explicit 3D Gaussian Splatting representation
~\cite{Kerbl2023GaussianSplatting}, where a scene is represented as a set
of anisotropic Gaussian primitives
\begin{linenomath}
\begin{equation}
G={(\mu_j,\Sigma_j,\alpha_j,c_j)}_{j=1}^{M}.
\end{equation}
\end{linenomath}
Here $\mu_j$ and $\Sigma_j$ denote the Gaussian center and covariance,
$\alpha_j$ is opacity, and $c_j$ represents view-dependent color
parameters. For each camera view $i$, differentiable splatting renders an
RGB image $\hat I_i$ from $G$.

The photo-only reconstruction optimizes the Gaussian scene using RGB
supervision,
\begin{linenomath}
\begin{equation}
R=\rho_r(\hat I_i,I_i).
\end{equation}
\end{linenomath}
Here $\rho_r$ combines the $\ell_1$ and D-SSIM photometric terms of
standard 3DGS training~\cite{Kerbl2023GaussianSplatting}. This objective
preserves video-driven appearance, but it does not directly constrain
facade surface geometry. In close-range urban captures, glass,
reflections, repeated windows, weak texture, and sparse SfM support can
therefore lead to visually plausible renderings with unstable surface
orientation. This is consistent with recent analyses of Gaussian
reconstruction, where redundant primitives can fit training views while
the underlying geometry remains weak, and density control can depend
strongly on the quality of SfM initialization~\cite{Lu2024ScaffoldGS,Zhang2024PixelGS}.

The MONO baseline follows the geometry-consistent component of the
Vid2Sim-style pipeline by using image-derived monocular depth and normal
cues during Gaussian optimization. These cues are useful but are still
inferred from ambiguous RGB evidence. Our method keeps the same
video-driven Gaussian backbone, but adds the exterior facade shell as a
stronger source of per-view geometric supervision.

\subsection{Facade Structural Shell Preparation and Alignment}

The facade shell $S$ provides exterior structural geometry of the visible
building facade. Such exterior geometry can be prepared from scan-based
building modeling, mobile mapping, photogrammetric reconstruction, or
other architectural-prior
pipelines~\cite{Rashdi2022ScanToBIM,Vega2024BIMCaP,Zhang2025GS4Buildings}.
Unlike a complete semantic building model, the shell excludes sensitive
or unnecessary information such as interiors, MEP systems, equipment
identifiers, and project metadata. The method uses only facade-level
geometry and the rendered per-view supervision.

\subsection{Shell-to-Camera Geometry Rendering}
\label{sec:method_rendering}

After alignment, the facade shell is rendered into the training camera
views. For each pixel $p$ in view $i$, we cast a camera ray through
$K_i$, $R_i$, and $t_i$, and query its first valid intersection with the
aligned shell $S^c$. We implement this ray-casting step using Open3D
RaycastingScene, which supports ray intersections with triangle meshes
and returns first-hit distances and primitive normals
~\cite{Open3DRaycastingScene}. This produces three per-view maps:
\begin{linenomath}
\begin{equation}
(Z_i^S,N_i^S,M_i^S)=R_S(S^c,K_i,R_i,t_i).
\end{equation}
\end{linenomath}
Here $Z_i^S$ is the forward shell depth, $N_i^S$ is the camera-space
shell normal, and $M_i^S\in\{0,1\}$ is a valid mask indicating pixels
where the ray intersects the shell.

\subsection{Shell-supervised Gaussian Optimization}
\label{sec:method_optimization}

During training, the Gaussian scene renders RGB, inverse depth, and
normal predictions for each view:
\begin{linenomath}
\begin{equation}
(\hat I_i,\hat D_i,\hat N_i)=R_G(G,K_i,R_i,t_i).
\end{equation}
\end{linenomath}
The shell-guided training objective used in the main comparison is
\begin{linenomath}
\begin{equation}
L=R+\alpha(t)D+\beta(t)N+\gamma(t)C.
\end{equation}
\end{linenomath}
Here $R$ is the RGB photometric loss, $D$ is the shell depth loss, $N$ is
the shell normal loss, and $C$ is an auxiliary geometry-consistency
regularizer inherited from the Vid2Sim-style Gaussian optimization
pipeline~\cite{Xie2025Vid2Sim}. The scalar weights $\alpha(t)$,
$\beta(t)$, and $\gamma(t)$ vary with the training iteration. The RGB
term preserves video-driven appearance, while the
shell-rendered depth and normal terms provide the main external geometry
supervision. This follows a broader line of work showing that geometric
cues such as depth and normals can regularize neural reconstruction
~\cite{Deng2022DSNeRF,Yu2022MonoSDF} and Gaussian Splatting geometry
~\cite{Turkulainen2025DNSplatter}. In contrast to image-derived or
sensor-derived priors, our supervision is rendered from an aligned
exterior facade shell.

The depth term operates on inverse depth. Let $B_i^S=1/Z_i^S$ denote the
shell inverse depth. We apply a local patch-normalized cross-correlation
objective
\begin{linenomath}
\begin{equation}
D=
\frac{1}{|\Omega_i^S|}
\sum_{p\in\Omega_i^S}
\left(1-\mathrm{NCC}(\hat D_i,B_i^S;P(p))\right).
\end{equation}
\end{linenomath}
Here $\Omega_i^S={p\mid M_i^S(p)=1}$ is the set of valid
shell-supported pixels and $P(p)$ denotes the local patch centered at
$p$ on which the NCC is evaluated. This inverse-depth NCC term acts as a
local structural cue rather than an absolute metric L1 depth objective.

The normal term encourages the rendered Gaussian normals to follow the
shell-rendered facade orientation:
\begin{linenomath}
\begin{equation}
N=
\frac{1}{|\Omega_i^S|}
\sum_{p\in\Omega_i^S}
\left(1-\left\langle \hat N_i(p),N_i^S(p)\right\rangle\right).
\end{equation}
\end{linenomath}
Both predicted and shell-rendered normals are unit-normalized and
represented in the camera coordinate frame. This term is especially
important for facade scenes, where repeated windows and reflective glass
can produce visually plausible but geometrically inconsistent surfaces.

The auxiliary term $C$ follows the geometry-consistency regularization
used in the Vid2Sim-style reconstruction backbone~\cite{Xie2025Vid2Sim}.
In our setting, it is evaluated with the same valid-mask gating as the
shell losses so that geometric regularization is concentrated on regions
supported by the rendered facade shell. Our ablation separates the
depth, normal, and auxiliary regularization terms; the results show that
the dominant improvement comes from the shell-rendered normal and depth
supervision rather than from the auxiliary regularizer alone.

\subsection{Shell-supervised Optimization Procedure}

Algorithm~\ref{alg:shell_supervision} summarizes the shell-supervised
optimization procedure. The shell is used as supervision rather than
initialization, preserving the video-derived appearance model while
regularizing only visible shell-supported facade regions.

\begin{algorithm}[t]
\caption{Shell-supervised optimization}
\label{alg:shell_supervision}
\footnotesize
\KwIn{RGB frames $I_i$ and cameras $C_i$ for $i=1,\ldots,N$, facade shell $S$}
\KwOut{optimized Gaussian scene $G$}
\KwInit{$G\leftarrow$ video-to-3D Gaussian initialization}
$S^c\leftarrow T_{S\rightarrow c}S$ \tcp*{fixed alignment}
\For{view $i=1,\ldots,N$}{
$(Z_i^S,N_i^S,M_i^S)\leftarrow R_S(S^c,C_i)$;
$\Omega_i^S\leftarrow$ valid pixels where $M_i^S(p)=1$;
}
\For{$t=1,\ldots,T$}{
sample view $i$;
$(\hat I_i,\hat D_i,\hat N_i)\leftarrow R_G(G,C_i)$;
$Q\leftarrow\alpha(t)D+\beta(t)N+\gamma(t)C$ on valid pixels $\Omega_i^S$;
$L\leftarrow R+Q$;
update $G$ using $\nabla_G L$ with densification and pruning while $t\le t_{\mathrm{dens}}$;
}
\end{algorithm}

\section{Experiments}

\subsection{Scenes and Data}

We evaluate our method on two anonymized close-range urban facade
sequences, denoted as the primary scene and the secondary scene. Each
sequence contains handheld monocular RGB frames, Structure-from-Motion
camera parameters, an anonymized exterior facade shell point cloud, a
fixed shell-to-camera alignment, and per-view shell-rendered depth,
normal, and valid-mask supervision. The camera parameters are estimated
using an SfM / Vid2Sim-style video-to-3D reconstruction
pipeline~\cite{Schoenberger2016SFM,Xie2025Vid2Sim}.
Table~\ref{tab:scene_stats} summarizes the scene statistics.

\begin{figure}[t]
\centering
\includegraphics[width=\columnwidth]{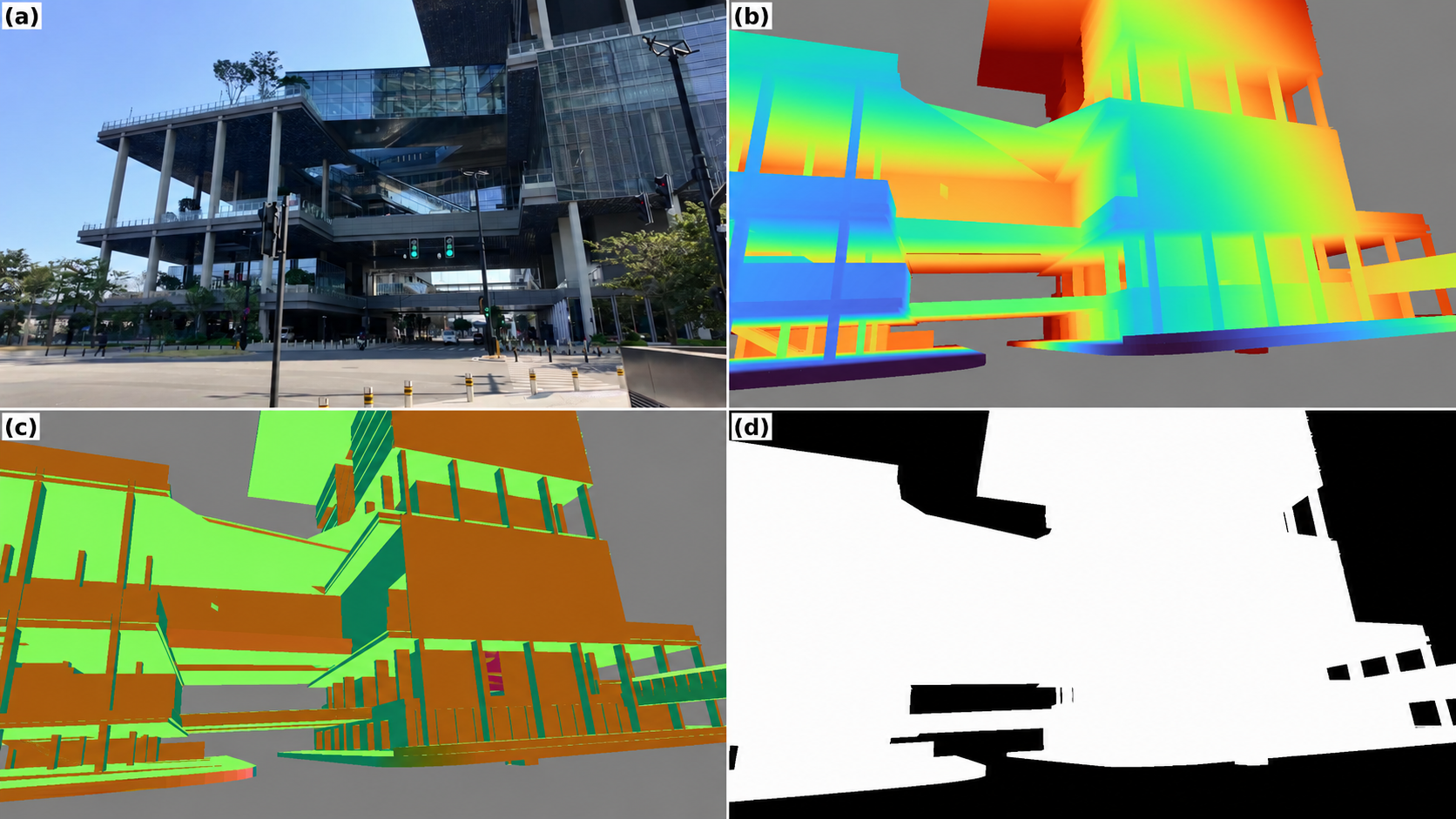}
\vspace{-0.5em}
\caption{Representative shell supervision on the primary scene: (a) RGB frame, (b) shell depth, (c) camera-space shell normal, and (d) valid mask. White pixels in the valid mask indicate shell-supported regions used for geometric supervision.}
\label{fig:primary_supervision}
\end{figure}

The primary scene is used for the full quantitative comparison and ablation study, while the secondary scene is used as a validation case for the same shell-supervision design on another facade geometry.
For both scenes, the facade shell is used only to produce
per-view geometric supervision, including shell depth,
camera-space shell normal, and a valid mask indicating pixels
where the shell is visible from the current camera.
Figure~\ref{fig:primary_supervision} shows a representative supervision example, and additional secondary-scene examples are provided in the supplementary material. All scene names, site identifiers, and project-specific metadata are anonymized in the released material.

\begin{table}[!t]
\centering
\scriptsize
\setlength{\tabcolsep}{2.8pt}
\resizebox{\columnwidth}{!}{
\begin{tabular}{lccccc}
\toprule
Scene & RGB & Tr./Te. & Res. & Shell & Role \\
\midrule
Primary & 201 & 175/26 & $1898{\times}1066$ & 120 & Main+abl. \\
Secondary & 52 & 45/7 & $1861{\times}1044$ & 52 & Val. \\
\bottomrule
\end{tabular}
}
\vspace{-0.4em}
\caption{
Scene statistics. ``Shell'' denotes views with rendered facade-shell
depth, normal, and valid-mask supervision.
}
\label{tab:scene_stats}
\vspace{-0.8em}
\end{table}

\subsection{Baselines}

We compare the proposed shell-guided reconstruction against both appearance-driven and geometry-aware Gaussian reconstruction baselines. All methods use the same RGB frames, camera parameters, train/test split, and initialization for each scene unless otherwise specified.

\textbf{Photo-only 3DGS.}
This baseline optimizes a standard 3D Gaussian representation using RGB photometric supervision only~\cite{Kerbl2023GaussianSplatting}. It represents the appearance-driven reconstruction setting and tests whether photometric training alone can recover reliable facade geometry in close-range urban scenes.

\textbf{MONO.}
This baseline follows the monocular-cue geometric supervision used in
our Vid2Sim-style reconstruction pipeline~\cite{Xie2025Vid2Sim}. It uses
image-derived depth and normal cues during Gaussian optimization, but
does not use the external facade shell. We treat MONO as a controlled
baseline for the geometry-consistent reconstruction component of a
real-to-sim pipeline, rather than as a complete reproduction of the full
Vid2Sim system, whose scope also includes hybrid simulation
construction, scene augmentation, and downstream embodied navigation.
This comparison tests whether a facade shell can provide stronger
structural supervision than monocular cues alone in reflective and
repetitive facade regions.

\textbf{2DGS.}
We include 2D Gaussian Splatting as a surface-oriented Gaussian baseline~\cite{Huang2024TwoDGS}. Unlike 3D Gaussian primitives, 2DGS represents scenes using oriented planar Gaussian disks and is designed to improve geometric consistency. This baseline tests whether a geometry-aware Gaussian representation alone can resolve the facade geometry ambiguity without external shell supervision.

\textbf{Ours.}
Our shell-guided model uses the same video-driven Gaussian reconstruction
framework, but adds mask-gated shell-rendered supervision. Specifically,
it combines RGB photometric supervision with inverse-depth NCC,
camera-space normal supervision, and an auxiliary geometry-consistency
regularizer on valid shell-supported pixels. This comparison isolates
the effect of using an aligned exterior facade shell as an external
geometric prior, while keeping the video-driven Gaussian backbone fixed.

\textbf{Ablation variants.}
On the primary scene, we evaluate three variants
that remove individual shell-guided terms: depth-only, normal-only, and
depth+normal without the auxiliary geometry-consistency regularizer.
These variants isolate surface-distance guidance, facade-orientation
guidance, and the effect of the auxiliary regularizer, following common
evaluations of depth and normal priors in Gaussian reconstruction
methods~\cite{Turkulainen2025DNSplatter}.

\subsection{Implementation Details}

All methods use the same RGB frames, SfM camera parameters, train/test split, image resolution, and sparse initialization within each scene unless otherwise specified. The photo-only, MONO, and shell-guided variants share the same 3DGS-style initialization produced by the SfM / Vid2Sim-style reconstruction pipeline~\cite{Schoenberger2016SFM,Xie2025Vid2Sim}. The 2DGS baseline is trained with the same camera split and image resolution, and is included as a surface-oriented Gaussian reconstruction baseline~\cite{Huang2024TwoDGS}.

For the shell-guided models, the facade shell is first aligned to the reconstruction coordinate frame and then kept fixed throughout training. We pre-render the aligned shell into each training view to obtain depth, camera-space normal, and valid-mask maps. These maps are not denoised or manually edited. During optimization, shell losses are evaluated only on valid shell-supported pixels. This mask-gated design prevents the shell prior from constraining sky, dynamic objects, occluded regions, or regions outside the available facade shell.

All methods are trained for 30k iterations following
the 3DGS-style optimization schedule~\cite{Kerbl2023GaussianSplatting}.
For the shell-supervised model, densification follows the 3DGS-style
training schedule and is stopped at 15k iterations. The weight of the
mask-gated inverse-depth NCC term is exponentially decayed from 0.5 to
0.1, while the normal and auxiliary geometry-consistency weights are
fixed at 0.1. This follows the broader
practice of using depth and normal cues to regularize Gaussian
reconstruction~\cite{Turkulainen2025DNSplatter} and the
geometry-supervised reconstruction setting of Vid2Sim~\cite{Xie2025Vid2Sim}.
The depth term used in our experiments is the inverse-depth NCC term in
Sec.~3.4, not a metric L1 depth loss.

For the primary-scene ablations, we keep the same training protocol and change only the shell-guided loss terms. The depth-only model keeps the shell depth term, the normal-only model keeps the shell normal term, and the depth+normal model removes the auxiliary geometry-consistency regularizer while keeping both shell-rendered cues. These variants follow common evaluations of depth and normal priors in Gaussian reconstruction~\cite{Turkulainen2025DNSplatter}.
Code will be released upon acceptance.

\subsection{Evaluation Metrics}

We evaluate each method using both appearance and geometry metrics.
Appearance metrics are computed on the held-out test views and include
PSNR, SSIM~\cite{Wang2004SSIM}, and LPIPS~\cite{Zhang2018LPIPS}.
These metrics measure whether shell supervision preserves the
video-driven rendering quality. Since our goal is not to maximize
photometric quality alone, we report appearance metrics together with
geometric measurements rather than using them as the sole criterion.

Because our target setting is urban real-to-sim reconstruction, we evaluate geometry in addition to appearance. We do not evaluate downstream reinforcement learning or interactive simulation in this paper; instead, we measure reconstruction properties that are needed before collision-aware simulation can be reliable, including shell-supported facade orientation and visible-surface consistency. This evaluation focus is consistent with embodied-simulation and real-to-sim settings, where reconstructed environments must support not only visual observations but also geometry for physical interaction, collision reasoning, and agent-environment contact~\cite{Dosovitskiy2017CARLA,Savva2019Habitat,Shen2021iGibson,Xie2025Vid2Sim}.

For image-space geometry evaluation, we compare rendered geometry
against the aligned facade-shell supervision on shell-supported pixels.
We report scale-aligned depth error and unsigned normal angular error.
For depth, rendered and shell depths are first converted to the same
metric coordinate scale, and the rendered depth is then aligned to the
shell depth by a per-view median scale factor computed only on valid
shell-supported pixels. We report the resulting D-MAE as a diagnostic
of shell-depth structural consistency, rather than as evidence of
absolute metric-depth recovery. This choice matches our inverse-depth
NCC training term, which encourages local depth-structure agreement but
does not directly impose a metric L1 depth objective.

For normals, we report the mean unsigned angular deviation between
rendered normals and shell-rendered camera-space normals on valid
shell-supported pixels. The angle is computed from the absolute dot
product between the two unit normals. This metric evaluates local facade
orientation rather than the normal-facing direction, which is the
geometric property most relevant for our shell-supported reconstruction diagnostic.

We further evaluate visible-surface point clouds using symmetric
Chamfer distance and thresholded F-score. For each evaluated view, we
back-project valid rendered depth pixels into 3D and compare the
reconstructed visible-surface points with the corresponding
shell-supported reference points. Chamfer distance measures the average
bidirectional nearest-neighbor distance between the two point sets.
F-score combines precision and recall under distance thresholds,
following common 3D reconstruction evaluation practice
~\cite{Knapitsch2017TanksTemples}. We report F-scores at multiple
thresholds to show whether a method improves both strict local accuracy
and coarse facade-level alignment.

All geometry metrics are computed only on valid shell-supported
regions. This avoids penalizing methods for sky, dynamic objects,
vegetation, occluded areas, or facade regions outside the available
exterior shell. On the primary scene, we use the same metrics for the
main comparison and the ablation study. On the secondary scene, we use
the same evaluation protocol to test whether the shell-guided
supervision transfers to another facade geometry.

\section{Results and Ablation}

% Figure 3: single-column qualitative geometry strip.
\begin{figure*}[!t]
\centering
\includegraphics[width=\textwidth]{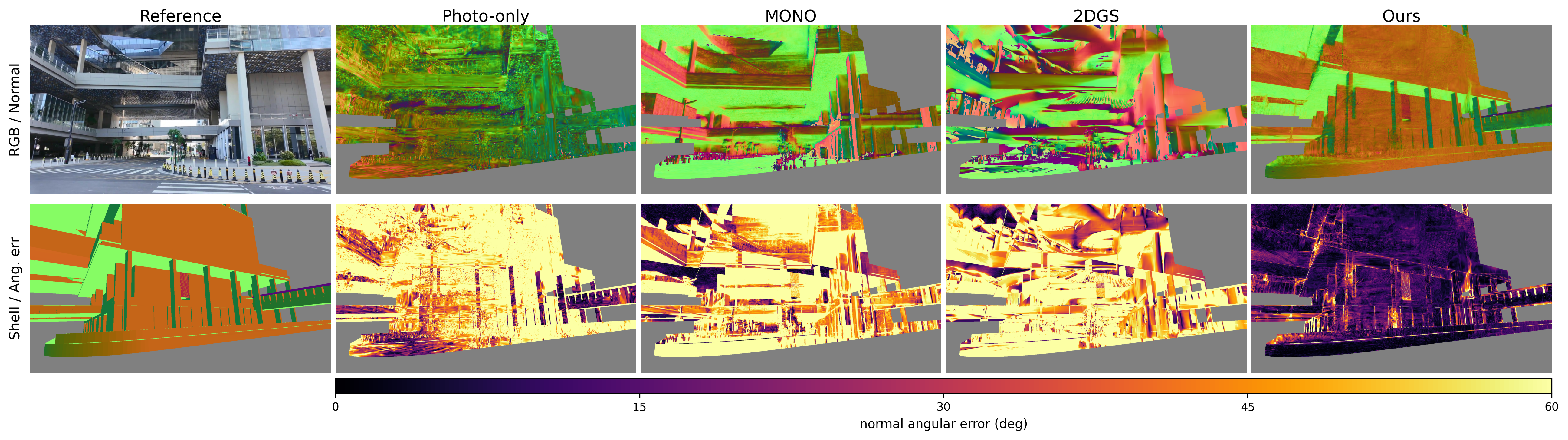}
\vspace{-1.2em}
\caption{Qualitative geometry comparison on a held-out test view of the primary scene. The first column provides the RGB context and shell-rendered reference, and the remaining columns show method outputs. The top row compares camera-space normal maps, while the bottom row shows unsigned angular normal error against the shell-rendered reference, with gray indicating pixels outside the valid shell mask. Photo-only, MONO, and 2DGS produce fragmented facade orientations, while shell-guided supervision yields more coherent shell-supported facade geometry.}
\label{fig:qualitative}
\vspace{-0.6em}
\end{figure*}

\begin{table}[!t]
\centering
\scriptsize
\setlength{\tabcolsep}{2.2pt}
\resizebox{\columnwidth}{!}{
\begin{tabular}{lccccccc}
\toprule
Method & PSNR$\uparrow$ & SSIM$\uparrow$ & LPIPS$\downarrow$ & D-MAE$\downarrow$ & Normal$\downarrow$ & $<10^\circ\uparrow$ & $<20^\circ\uparrow$ \\
\midrule
Photo-only~\cite{Kerbl2023GaussianSplatting} & 23.67 & 0.835 & 0.262 & 0.236 & 55.21 & 2.30 & 7.71 \\
MONO~\cite{Xie2025Vid2Sim} & 23.48 & 0.832 & 0.269 & 0.220 & 53.57 & 7.22 & 14.65 \\
2DGS~\cite{Huang2024TwoDGS} & 23.16 & 0.824 & 0.303 & \textbf{0.201} & 57.46 & 3.95 & 10.76 \\
Ours & 23.06 & 0.824 & 0.282 & 0.273 & \textbf{17.42} & \textbf{47.36} & \textbf{73.12} \\
\bottomrule
\end{tabular}}
\vspace{-0.4em}
\caption{
Main quantitative results on the primary scene.
Appearance metrics are computed on held-out test views;
geometry metrics are computed on shell-supported test views.
D-MAE is per-view median-scale-aligned depth MAE in meters and is used as a depth diagnostic.
Normal is unsigned mean angular error in degrees.
}
\label{tab:main_results}
\vspace{-0.8em}
\end{table}

\subsection{Main Quantitative Results}

Table~\ref{tab:main_results} reports the main results on the primary scene. The photo-only 3DGS baseline achieves the best photometric scores, which is expected because it optimizes only the RGB reconstruction objective~\cite{Kerbl2023GaussianSplatting}. The proposed shell-guided model slightly reduces PSNR compared with photo-only 3DGS, from 23.67 dB to 23.06 dB, and gives similar SSIM while maintaining a lower LPIPS than 2DGS~\cite{Huang2024TwoDGS}. This indicates that the shell losses do not aim to maximize photometric quality alone, but preserve the video-driven appearance to a comparable level.

The strongest benefit appears in facade orientation. Photo-only 3DGS, MONO~\cite{Xie2025Vid2Sim}, and 2DGS all produce large unsigned normal errors above 53 degrees on shell-supported test views, showing that photometric reconstruction, monocular cues, and a surface-oriented Gaussian representation alone are insufficient to recover consistent facade orientation in this scene. In contrast, our shell-guided model reduces the mean normal error to 17.42 degrees and increases the percentage of pixels below 20 degrees from 7.71\% for photo-only 3DGS and 14.65\% for MONO to 73.12\%.

D-MAE is a scale-aligned depth diagnostic, not a measure of absolute metric-depth accuracy. Our depth term uses local inverse-depth NCC, which encourages structural depth agreement but does not impose a metric L1 objective. Consistent with this, 2DGS obtains the lowest D-MAE while its normal error remains high. Our method has a larger D-MAE but substantially better facade orientation.

This apparent discrepancy arises because D-MAE and the visible-surface point-cloud metrics evaluate different properties: D-MAE applies per-view median depth rescaling and measures scale-normalized image-space depth agreement, whereas Table~\ref{tab:pc_chamfer} uses raw rendered depth for 3D back-projection without per-view rescaling. Therefore, D-MAE diagnoses local depth-structure agreement, while the point-cloud evaluation reflects raw visible-surface consistency.

\subsection{Visible-surface Point-cloud Evaluation}

The image-space geometry metrics in Table~\ref{tab:main_results} show that shell guidance substantially improves facade normal orientation. We further evaluate whether this improvement is reflected in the reconstructed visible 3D surface. For each shell-supported held-out test view, we back-project valid rendered depth pixels into 3D and compare the resulting visible-surface point cloud with the corresponding shell-supported reference points. We report symmetric Chamfer distance and thresholded F-scores following common 3D reconstruction evaluation practice~\cite{Knapitsch2017TanksTemples}. Unlike D-MAE, this point-cloud evaluation uses raw rendered depth for back-projection and does not apply per-view depth rescaling.

\begin{table}[t]
\centering
\small
\setlength{\tabcolsep}{0pt}
\begin{tabular*}{\columnwidth}{@{\extracolsep{\fill}}lcccccc@{}}
\toprule
Method & Acc.$\downarrow$ & Comp.$\downarrow$ & CD$\downarrow$ & F$_5$$\uparrow$ & F$_{10}$$\uparrow$ & F$_{20}$$\uparrow$ \\
\midrule
Photo-only~\cite{Kerbl2023GaussianSplatting} & 0.091 & 0.048 & 0.069 & 0.546 & 0.779 & 0.921 \\
MONO~\cite{Xie2025Vid2Sim}              & 0.090 & 0.052 & 0.071 & 0.547 & 0.767 & 0.916 \\
2DGS~\cite{Huang2024TwoDGS}             & 0.077 & 0.055 & 0.066 & 0.563 & 0.788 & \textbf{0.938} \\
Ours                                    & \textbf{0.069} & \textbf{0.036} & \textbf{0.052} & \textbf{0.710} & \textbf{0.865} & 0.937 \\
\bottomrule
\end{tabular*}
\caption{Visible-surface point-cloud evaluation on the primary scene.
All metrics are computed on shell-supported held-out test views using
raw rendered depth without per-view depth rescaling for point-cloud
back-projection. Distances are reported in meters. F$_5$, F$_{10}$, and
F$_{20}$ denote F-scores at 5\,cm, 10\,cm, and 20\,cm thresholds.}
\label{tab:pc_chamfer}
\end{table}

Table~\ref{tab:pc_chamfer} shows that the proposed shell-guided model improves visible-surface geometry over both appearance-driven and geometry-aware baselines. Compared with photo-only 3DGS~\cite{Kerbl2023GaussianSplatting}, our method reduces Chamfer distance from 0.069~m to 0.052~m and improves F$_5$ from 0.546 to 0.710. The MONO baseline~\cite{Xie2025Vid2Sim} produces similar point-cloud accuracy to photo-only reconstruction, indicating that image-derived monocular cues alone do not reliably correct the facade surface in this scene.

The 2DGS baseline~\cite{Huang2024TwoDGS} modestly improves Chamfer distance over photo-only 3DGS, from 0.069~m to 0.066~m, but remains substantially worse than the shell-guided model at stricter thresholds. In particular, Ours improves F$_5$ from 0.563 to 0.710 and F$_{10}$ from 0.788 to 0.865 relative to 2DGS, while both methods are comparable at the loose 20~cm threshold. This suggests that changing the Gaussian primitive representation improves coarse visible-surface consistency, but external facade-shell supervision is more effective for recovering accurate shell-supported facade geometry.

This evaluation remains restricted to visible shell-supported regions. It should therefore be interpreted as visible facade-shell consistency, not as evidence that the complete scene or an extracted mesh is geometrically perfect.

\subsection{Ablation of Shell Supervision}
\label{sec:ablation}

\begin{table}[t]
\centering
\small
\setlength{\tabcolsep}{3.5pt}
\renewcommand{\arraystretch}{1.05}
\begin{tabular}{lccc}
\toprule
Variant & D-MAE$\downarrow$ & Normal$\downarrow$ & $<20^\circ\uparrow$ \\
\midrule
Photo-only~\cite{Kerbl2023GaussianSplatting} & 0.236 & 55.21 & 7.71 \\
Depth only & \textbf{0.113} & 55.42 & 6.74 \\
Normal only & 0.183 & 17.74 & 72.33 \\
Depth+normal & 0.231 & \textbf{17.09} & \textbf{73.80} \\
Full objective & 0.273 & 17.42 & 73.12 \\
\bottomrule
\end{tabular}
\vspace{-0.3em}
\caption{Ablation of shell-guided supervision terms on the primary scene.
D-MAE is the median-scale-aligned depth diagnostic. Normal denotes
unsigned mean angular error in degrees. The depth term improves the
depth diagnostic, while shell-rendered normal supervision provides the
dominant facade-orientation gain.}
\label{tab:ablation_main}
\vspace{-0.6em}
\end{table}

Table~\ref{tab:ablation_main} ablates the shell-guided loss terms on the primary scene, following common evaluations of depth and normal priors in Gaussian reconstruction~\cite{Turkulainen2025DNSplatter}. The depth-only variant substantially improves the scale-aligned depth diagnostic, but does not improve facade orientation. In contrast, shell-rendered normal supervision is the dominant cue for orientation consistency, reducing mean normal error from 55.21 degrees for photo-only 3DGS~\cite{Kerbl2023GaussianSplatting} to 17.74 degrees. Combining depth and normal terms gives the best image-space orientation metrics, indicating that shell depth provides complementary surface-distance guidance.

The full objective obtains comparable orientation performance while retaining the auxiliary geometry-consistency regularizer used by the underlying Vid2Sim-style reconstruction backbone~\cite{Xie2025Vid2Sim}. These results show that the main improvement comes from shell-rendered depth and normal supervision rather than from the auxiliary regularizer alone. We therefore use the full objective as the default model in the main comparison. The complete supplementary ablation additionally reports the $<10^\circ$ normal-accuracy metric.

\subsection{Qualitative Analysis and Cross-scene Validation}
\label{sec:cross_scene_qualitative}

Figure~\ref{fig:qualitative} compares rendered facade geometry on a held-out test view of the primary scene. The photo-only and MONO reconstructions produce fragmented normal fields on the glass curtain wall and repeated-window regions: surface orientation varies almost pixel-to-pixel even where the underlying facade is planar. 2DGS, despite its surface-oriented representation, also does not recover consistent facade orientation in this close-range setting. The shell-guided model also follows the shell-rendered reference while retaining fine structures such as columns and slab edges.

We further evaluate the same supervision design on the secondary scene, which contains fewer frames and noisier shell-supported regions than the primary scene; we therefore use it as a cross-scene validation case rather than a large-scale generalization benchmark. With the same training protocol and no manual editing of the shell-rendered supervision maps, the secondary scene follows the same overall trend: shell-guided supervision improves facade-orientation consistency on valid shell-supported pixels while preserving comparable rendering quality. Additional quantitative results and supervision examples for the secondary scene are provided in the supplementary material.

\section{Discussion and Limitations}

The proposed method improves the reconstruction stage of a
Vid2Sim-style real-to-sim pipeline, rather than providing a complete
embodied simulation system. Accordingly, our results should be
interpreted as asset-level geometry improvements, not as downstream
simulation or navigation validation
~\cite{Xie2025Vid2Sim,Savva2019Habitat,Shen2021iGibson,Dosovitskiy2017CARLA}.

A key design choice is that the method uses only an exterior structural
shell, rather than a complete semantic building model. Such shells can
come from existing exterior models, terrestrial or mobile laser scanning,
aerial or street-level mapping, or photogrammetric reconstruction~\cite{Rashdi2022ScanToBIM}.
The shell is converted into per-view depth, normal, and valid-mask
supervision, preserving video-driven appearance quality while improving
visible facade geometry where photometric and monocular cues are
ambiguous. The ablation shows that shell-rendered normal supervision is
the dominant cue for facade orientation, while shell depth provides
complementary structural guidance; the auxiliary regularizer is not the
main source of the improvement.

The method has several limitations. It assumes that an exterior shell can
be prepared and aligned to the video reconstruction frame, and we do not
treat this alignment as an absolute metric reference. Supervision applies
only to shell-supported facade regions and does not explain sky,
vegetation, dynamic objects, interiors, or regions outside the shell. The
spatial-region holdout in the supplementary material shows that
extrapolation to disjoint unsupervised facade bands remains limited.
Glass also remains difficult: our current formulation does not model
transparency, reflection, refraction, or material appearance, and should
not be interpreted as solving glass facade reconstruction~\cite{Whelan2018MirrorGlass}.
Finally, improved visible geometry is only a prerequisite for downstream
interactive simulation; we do not claim a complete clean mesh, a finished
interactive simulation environment, or downstream navigation or
reinforcement-learning results.

\section{Conclusion}
We presented a facade shell-guided Gaussian reconstruction framework for
the video-to-3D stage of urban real-to-sim pipelines. The method renders
an aligned exterior shell into per-view depth, camera-space normal, and
valid-mask supervision for mask-gated Gaussian optimization. Experiments on close-range urban facade scenes show improved visible
facade orientation and visible-surface point-cloud consistency in
raw-depth visible-surface point-cloud evaluation, while preserving
comparable RGB rendering quality. These results support
exterior facade shells as practical structural priors for improving
shell-supported geometry in video-driven urban facade reconstruction.

{
    \small
    \bibliographystyle{ieeenat_fullname}
    \bibliography{main}

@article{Kerbl2023GaussianSplatting,
  author  = {Kerbl, Bernhard and Kopanas, Georgios and Leimk{\"u}hler, Thomas and Drettakis, George},
  title   = {{3D} Gaussian Splatting for Real-Time Radiance Field Rendering},
  journal = {ACM Transactions on Graphics},
  volume  = {42},
  number  = {4},
  pages   = {1--14},
  year    = {2023}
}

@inproceedings{Schoenberger2016SFM,
title     = {Structure-from-Motion Revisited},
author    = {Sch{\"o}nberger, Johannes L. and Frahm, Jan-Michael},
booktitle = {Proceedings of the IEEE Conference on Computer Vision and Pattern Recognition (CVPR)},
pages     = {4104--4113},
year      = {2016}
}

@inproceedings{Xie2025Vid2Sim,
title     = {Vid2Sim: Realistic and Interactive Simulation from Video for Urban Navigation},
author    = {Xie, Ziyang and Liu, Zhizheng and Peng, Zhenghao and Wu, Wayne and Zhou, Bolei},
booktitle = {Proceedings of the IEEE/CVF Conference on Computer Vision and Pattern Recognition (CVPR)},
pages     = {1581--1591},
year      = {2025}
}

@inproceedings{Turkulainen2025DNSplatter,
title     = {{DN}-Splatter: Depth and Normal Priors for Gaussian Splatting and Meshing},
author    = {Turkulainen, Matias and Ren, Xuqian and Melekhov, Iaroslav and Seiskari, Otto and Rahtu, Esa and Kannala, Juho},
booktitle = {Proceedings of the IEEE/CVF Winter Conference on Applications of Computer Vision (WACV)},
pages     = {2421--2431},
year      = {2025}
}

@inproceedings{Huang2024TwoDGS,
title     = {{2D} Gaussian Splatting for Geometrically Accurate Radiance Fields},
author    = {Huang, Binbin and Yu, Zehao and Chen, Anpei and Geiger, Andreas and Gao, Shenghua},
booktitle = {ACM SIGGRAPH 2024 Conference Papers},
pages     = {1--11},
year      = {2024}
}

@misc{Open3DRaycastingScene,
title        = {{Open3D} RaycastingScene Documentation},
author       = {{Open3D Developers}},
howpublished = {\url{https://www.open3d.org/docs/latest/python_api/open3d.t.geometry.RaycastingScene.html}},
note         = {Accessed: 2026-06-17},
year         = {2026}
}

@article{Zhang2025GS4Buildings,
title   = {{GS4Buildings}: Prior-Guided Gaussian Splatting for {3D} Building Reconstruction},
author  = {Zhang, Qilin and Wysocki, Olaf and Jutzi, Boris},
journal = {ISPRS Annals of the Photogrammetry, Remote Sensing and Spatial Information Sciences},
volume  = {X-4/W6-2025},
pages   = {249--256},
year    = {2025}
}

@article{Wang2004SSIM,
title   = {Image Quality Assessment: From Error Visibility to Structural Similarity},
author  = {Wang, Zhou and Bovik, Alan C. and Sheikh, Hamid R. and Simoncelli, Eero P.},
journal = {IEEE Transactions on Image Processing},
volume  = {13},
number  = {4},
pages   = {600--612},
year    = {2004}
}

@inproceedings{Zhang2018LPIPS,
title     = {The Unreasonable Effectiveness of Deep Features as a Perceptual Metric},
author    = {Zhang, Richard and Isola, Phillip and Efros, Alexei A. and Shechtman, Eli and Wang, Oliver},
booktitle = {Proceedings of the IEEE Conference on Computer Vision and Pattern Recognition (CVPR)},
pages     = {586--595},
year      = {2018}
}

@article{Knapitsch2017TanksTemples,
title     = {Tanks and Temples: Benchmarking Large-Scale Scene Reconstruction},
author    = {Knapitsch, Arno and Park, Jaesik and Zhou, Qian-Yi and Koltun, Vladlen},
journal   = {ACM Transactions on Graphics},
volume    = {36},
number    = {4},
pages     = {1--13},
year      = {2017}
}

@article{Rashdi2022ScanToBIM,
title   = {Scanning Technologies to Building Information Modelling: A Review},
author  = {Rashdi, Rabia and Mart{\'i}nez-S{\'a}nchez, Joaqu{\'i}n and Arias, Pedro and Qiu, Zhouyan},
journal = {Infrastructures},
volume  = {7},
number  = {4},
pages   = {49},
year    = {2022}
}

@article{Whelan2018MirrorGlass,
title     = {Reconstructing Scenes with Mirror and Glass Surfaces},
author    = {Whelan, Thomas and Goesele, Michael and Lovegrove, Steven J. and Straub, Julian and Green, Simon and Szeliski, Richard and Butterfield, Steven and Verma, Shobhit and Newcombe, Richard},
journal   = {ACM Transactions on Graphics},
volume    = {37},
number    = {4},
pages     = {1--11},
year      = {2018}
}

@inproceedings{Mildenhall2020NeRF,
title     = {{NeRF}: Representing Scenes as Neural Radiance Fields for View Synthesis},
author    = {Mildenhall, Ben and Srinivasan, Pratul P. and Tancik, Matthew and Barron, Jonathan T. and Ramamoorthi, Ravi and Ng, Ren},
booktitle = {Proceedings of the European Conference on Computer Vision (ECCV)},
pages     = {405--421},
year      = {2020}
}

@inproceedings{Deng2022DSNeRF,
title     = {Depth-supervised {NeRF}: Fewer Views and Faster Training for Free},
author    = {Deng, Kangle and Liu, Andrew and Zhu, Jun-Yan and Ramanan, Deva},
booktitle = {Proceedings of the IEEE/CVF Conference on Computer Vision and Pattern Recognition (CVPR)},
pages     = {12882--12891},
year      = {2022}
}

@inproceedings{Yu2022MonoSDF,
title     = {{MonoSDF}: Exploring Monocular Geometric Cues for Neural Implicit Surface Reconstruction},
author    = {Yu, Zehao and Peng, Songyou and Niemeyer, Michael and Sattler, Torsten and Geiger, Andreas},
booktitle = {Advances in Neural Information Processing Systems (NeurIPS)},
volume    = {35},
year      = {2022}
}

@inproceedings{Barron2022MipNeRF360,
title     = {Mip-{NeRF} 360: Unbounded Anti-Aliased Neural Radiance Fields},
author    = {Barron, Jonathan T. and Mildenhall, Ben and Verbin, Dor and Srinivasan, Pratul P. and Hedman, Peter},
booktitle = {Proceedings of the IEEE/CVF Conference on Computer Vision and Pattern Recognition (CVPR)},
pages     = {5470--5479},
year      = {2022}
}

@article{Muller2022InstantNGP,
title     = {Instant Neural Graphics Primitives with a Multiresolution Hash Encoding},
author    = {M{\"u}ller, Thomas and Evans, Alex and Schied, Christoph and Keller, Alexander},
journal   = {ACM Transactions on Graphics},
volume    = {41},
number    = {4},
pages     = {1--15},
year      = {2022}
}

@inproceedings{Tancik2022BlockNeRF,
title     = {Block-{NeRF}: Scalable Large Scene Neural View Synthesis},
author    = {Tancik, Matthew and Casser, Vincent and Yan, Xinchen and Pradhan, Sabeek and Mildenhall, Ben and Srinivasan, Pratul P. and Barron, Jonathan T. and Kretzschmar, Henrik},
booktitle = {Proceedings of the IEEE/CVF Conference on Computer Vision and Pattern Recognition (CVPR)},
pages     = {8248--8258},
year      = {2022}
}

@inproceedings{Turki2022MegaNeRF,
title     = {Mega-{NeRF}: Scalable Construction of Large-Scale {NeRF}s for Virtual Fly-Throughs},
author    = {Turki, Haithem and Ramanan, Deva and Satyanarayanan, Mahadev},
booktitle = {Proceedings of the IEEE/CVF Conference on Computer Vision and Pattern Recognition (CVPR)},
pages     = {12922--12931},
year      = {2022}
}

@inproceedings{Rematas2022UrbanRF,
title     = {Urban Radiance Fields},
author    = {Rematas, Konstantinos and Liu, Andrew and Srinivasan, Pratul P. and Barron, Jonathan T. and Tagliasacchi, Andrea and Funkhouser, Thomas and Ferrari, Vittorio},
booktitle = {Proceedings of the IEEE/CVF Conference on Computer Vision and Pattern Recognition (CVPR)},
pages     = {12932--12942},
year      = {2022}
}

@inproceedings{Zanjani2025PlanarGS,
title     = {Planar Gaussian Splatting},
author    = {Zanjani, Farhad G. and Cai, Hong and Ackermann, Hanno and Mirvakhabova, Leila and Porikli, Fatih},
booktitle = {Proceedings of the IEEE/CVF Winter Conference on Applications of Computer Vision (WACV)},
pages     = {8887--8896},
year      = {2025}
}

@inproceedings{Guedon2024SuGaR,
title     = {{SuGaR}: Surface-Aligned Gaussian Splatting for Efficient {3D} Mesh Reconstruction and High-Quality Mesh Rendering},
author    = {Gu{\'e}don, Antoine and Lepetit, Vincent},
booktitle = {Proceedings of the IEEE/CVF Conference on Computer Vision and Pattern Recognition (CVPR)},
pages     = {5354--5363},
year      = {2024}
}

@article{Yu2024GOF,
title     = {Gaussian Opacity Fields: Efficient Adaptive Surface Reconstruction in Unbounded Scenes},
author    = {Yu, Zehao and Sattler, Torsten and Geiger, Andreas},
journal   = {ACM Transactions on Graphics},
volume    = {43},
number    = {6},
pages     = {1--13},
year      = {2024}
}

@inproceedings{Jiang2025GaussianSurfels,
title     = {Geometry Field Splatting with Gaussian Surfels},
author    = {Jiang, Kaiwen and Sivaram, Venkataram and Peng, Cheng and Ramamoorthi, Ravi},
booktitle = {Proceedings of the IEEE/CVF Conference on Computer Vision and Pattern Recognition (CVPR)},
pages     = {5752--5762},
year      = {2025}
}

@inproceedings{Yariv2021VolSDF,
title     = {Volume Rendering of Neural Implicit Surfaces},
author    = {Yariv, Lior and Gu, Jiatao and Kasten, Yoni and Lipman, Yaron},
booktitle = {Advances in Neural Information Processing Systems (NeurIPS)},
volume    = {34},
year      = {2021}
}

@inproceedings{Wang2021NeuS,
title     = {{NeuS}: Learning Neural Implicit Surfaces by Volume Rendering for Multi-view Reconstruction},
author    = {Wang, Peng and Liu, Lingjie and Liu, Yuan and Theobalt, Christian and Komura, Taku and Wang, Wenping},
booktitle = {Advances in Neural Information Processing Systems (NeurIPS)},
volume    = {34},
pages     = {27171--27183},
year      = {2021}
}

@inproceedings{Li2023Neuralangelo,
title     = {Neuralangelo: High-Fidelity Neural Surface Reconstruction},
author    = {Li, Zhaoshuo and M{\"u}ller, Thomas and Evans, Alex and Taylor, Russell H. and Unberath, Mathias and Liu, Ming-Yu and Lin, Chen-Hsuan},
booktitle = {Proceedings of the IEEE/CVF Conference on Computer Vision and Pattern Recognition (CVPR)},
pages     = {8456--8465},
year      = {2023}
}

@inproceedings{Niemeyer2022RegNeRF,
title     = {Reg{NeRF}: Regularizing Neural Radiance Fields for View Synthesis from Sparse Inputs},
author    = {Niemeyer, Michael and Barron, Jonathan T. and Mildenhall, Ben and Sajjadi, Mehdi S. M. and Geiger, Andreas and Radwan, Noha},
booktitle = {Proceedings of the IEEE/CVF Conference on Computer Vision and Pattern Recognition (CVPR)},
pages     = {5480--5490},
year      = {2022}
}

@inproceedings{Chang2017Matterport3D,
title     = {Matterport{3D}: Learning from {RGB-D} Data in Indoor Environments},
author    = {Chang, Angel and Dai, Angela and Funkhouser, Thomas and Halber, Maciej and Nie{\ss}ner, Matthias and Savva, Manolis and Song, Shuran and Zeng, Andy and Zhang, Yinda},
booktitle = {Proceedings of the International Conference on 3D Vision (3DV)},
pages     = {667--676},
year      = {2017}
}

@inproceedings{Xia2018GibsonEnv,
title     = {Gibson Env: Real-World Perception for Embodied Agents},
author    = {Xia, Fei and Zamir, Amir R. and He, Zhi-Yang and Sax, Alexander and Malik, Jitendra and Savarese, Silvio},
booktitle = {Proceedings of the IEEE Conference on Computer Vision and Pattern Recognition (CVPR)},
pages     = {9068--9079},
year      = {2018}
}

@inproceedings{Savva2019Habitat,
title     = {Habitat: A Platform for Embodied {AI} Research},
author    = {Savva, Manolis and Kadian, Abhishek and Maksymets, Oleksandr and Zhao, Yili and Wijmans, Erik and Jain, Bhavana and Straub, Julian and Liu, Jia and Koltun, Vladlen and Malik, Jitendra and Parikh, Devi and Batra, Dhruv},
booktitle = {Proceedings of the IEEE/CVF International Conference on Computer Vision (ICCV)},
pages     = {9339--9347},
year      = {2019}
}

@article{Kolve2017AI2THOR,
title         = {{AI2-THOR}: An Interactive {3D} Environment for Visual {AI}},
author        = {Kolve, Eric and Mottaghi, Roozbeh and Han, Winson and VanderBilt, Eli and Weihs, Luca and Herrasti, Alvaro and Deitke, Matt and Ehsani, Kiana and Gordon, Daniel and Zhu, Yuke and Kembhavi, Aniruddha and Gupta, Abhinav and Farhadi, Ali},
journal       = {arXiv preprint arXiv:1712.05474},
year          = {2017},
archivePrefix = {arXiv},
eprint        = {1712.05474}
}

@inproceedings{Shen2021iGibson,
title     = {iGibson 1.0: A Simulation Environment for Interactive Tasks in Large Realistic Scenes},
author    = {Shen, Bokui and Xia, Fei and Li, Chengshu and Mart{\'i}n-Mart{\'i}n, Roberto and Fan, Linxi and Wang, Guanzhi and P{\'e}rez-D'Arpino, Claudia and Buch, Shyamal and Srivastava, Sanjana and Tchapmi, Lyne and Tchapmi, Micael and Vainio, Kent and Wong, Josiah and Li, Fei-Fei and Savarese, Silvio},
booktitle = {Proceedings of the IEEE/RSJ International Conference on Intelligent Robots and Systems (IROS)},
pages     = {7520--7527},
year      = {2021}
}

@inproceedings{Dosovitskiy2017CARLA,
title     = {{CARLA}: An Open Urban Driving Simulator},
author    = {Dosovitskiy, Alexey and Ros, German and Codevilla, Felipe and L{\'o}pez, Antonio M. and Koltun, Vladlen},
booktitle = {Proceedings of the 1st Annual Conference on Robot Learning (CoRL)},
series    = {Proceedings of Machine Learning Research},
volume    = {78},
pages     = {1--16},
year      = {2017},
publisher = {PMLR}
}

@inproceedings{Xia2024Video2Game,
title     = {Video2Game: Real-time, Interactive, Realistic and Browser-Compatible Environment from a Single Video},
author    = {Xia, Hongchi and Lin, Zhi-Hao and Ma, Wei-Chiu and Wang, Shenlong},
booktitle = {Proceedings of the IEEE/CVF Conference on Computer Vision and Pattern Recognition (CVPR)},
pages     = {4578--4588},
year      = {2024}
}

@inproceedings{Vega2024BIMCaP,
title     = {{BIMCaP}: {BIM}-based {AI}-supported {LiDAR}-Camera Pose Refinement},
author    = {Vega-Torres, Miguel Arturo and Ribic, Anna and Garc{\'i}a de Soto, Borja and Borrmann, Andr{\'e}},
booktitle = {Proceedings of the 31st International Workshop on Intelligent Computing in Engineering (EG-ICE 2024)},
pages     = {423--432},
year      = {2024}
}

@inproceedings{Xie2023SNeRF,
title     = {{S-NeRF}: Neural Radiance Fields for Street Views},
author    = {Xie, Ziyang and Zhang, Junge and Li, Wenye and Zhang, Feihu and Zhang, Li},
booktitle = {International Conference on Learning Representations (ICLR)},
year      = {2023}
}

@article{Guo2023StreetSurf,
title         = {StreetSurf: Extending Multi-view Implicit Surface Reconstruction to Street Views},
author        = {Guo, Jianfei and Deng, Nianchen and Li, Xinyang and Bai, Yeqi and Shi, Botian and Wang, Chiyu and Ding, Chenjing and Wang, Dongliang and Li, Yikang},
journal       = {arXiv preprint arXiv:2306.04988},
year          = {2023},
archivePrefix = {arXiv},
eprint        = {2306.04988}
}

@inproceedings{Yan2024StreetGaussians,
title     = {Street Gaussians: Modeling Dynamic Urban Scenes with Gaussian Splatting},
author    = {Yan, Yunzhi and Lin, Haotong and Zhou, Chenxu and Wang, Weijie and Sun, Haiyang and Zhan, Kun and Lang, Xianpeng and Zhou, Xiaowei and Peng, Sida},
booktitle = {Proceedings of the European Conference on Computer Vision (ECCV)},
pages     = {156--173},
year      = {2024}
}

@inproceedings{Sun2021NeuralRecon,
title     = {NeuralRecon: Real-Time Coherent {3D} Reconstruction from Monocular Video},
author    = {Sun, Jiaming and Xie, Yiming and Chen, Linghao and Zhou, Xiaowei and Bao, Hujun},
booktitle = {Proceedings of the IEEE/CVF Conference on Computer Vision and Pattern Recognition (CVPR)},
pages     = {15598--15607},
year      = {2021}
}

@inproceedings{Guo2022ManhattanSDF,
title     = {Neural {3D} Scene Reconstruction with the Manhattan-world Assumption},
author    = {Guo, Haoyu and Peng, Sida and Lin, Haotong and Wang, Qianqian and Zhang, Guofeng and Bao, Hujun and Zhou, Xiaowei},
booktitle = {Proceedings of the IEEE/CVF Conference on Computer Vision and Pattern Recognition (CVPR)},
pages     = {5511--5520},
year      = {2022}
}

@inproceedings{Wang2022NeuRIS,
title     = {{NeuRIS}: Neural Reconstruction of Indoor Scenes Using Normal Priors},
author    = {Wang, Jiepeng and Wang, Peng and Long, Xiaoxiao and Theobalt, Christian and Komura, Taku and Liu, Lingjie and Wang, Wenping},
booktitle = {Proceedings of the European Conference on Computer Vision (ECCV)},
pages     = {139--155},
year      = {2022}
}

@inproceedings{Lu2024ScaffoldGS,
title     = {Scaffold-{GS}: Structured {3D} Gaussians for View-Adaptive Rendering},
author    = {Lu, Tao and Yu, Mulin and Xu, Linning and Xiangli, Yuanbo and Wang, Limin and Lin, Dahua and Dai, Bo},
booktitle = {Proceedings of the IEEE/CVF Conference on Computer Vision and Pattern Recognition (CVPR)},
pages     = {20654--20664},
year      = {2024}
}

@inproceedings{Zhang2024PixelGS,
title     = {Pixel-{GS}: Density Control with Pixel-Aware Gradient for {3D} Gaussian Splatting},
author    = {Zhang, Zheng and Hu, Wenbo and Lao, Yixing and He, Tong and Zhao, Hengshuang},
booktitle = {Proceedings of the European Conference on Computer Vision (ECCV)},
pages     = {326--342},
year      = {2024}
}

@inproceedings{Yu2024MipSplatting,
title     = {Mip-Splatting: Alias-Free {3D} Gaussian Splatting},
author    = {Yu, Zehao and Chen, Anpei and Huang, Binbin and Sattler, Torsten and Geiger, Andreas},
booktitle = {Proceedings of the IEEE/CVF Conference on Computer Vision and Pattern Recognition (CVPR)},
pages     = {19447--19456},
year      = {2024}
}
}

\end{document}

\title{Supplementary Material: Shell-Supervised Gaussian Splatting for Urban Real-to-Sim Reconstruction}

\author{
\begin{tabular}{c}
Yuan Yang$^{1}$ \quad
Peijun Lu$^{2}$ \quad
Fangzhou Lu$^{1}$ \quad
Sai Fan$^{1}$ \quad
Siqi Yan$^{1}$ \quad
Chenyuan Zhang$^{1}$ \\
Haobo Liang$^{1,\dagger}$ \quad
Yicheng Wang$^{1,\dagger}$ \\
{\small $^{1}$Hong Kong Center for Construction Robotics \quad
$^{2}$Tsinghua University} \\
{\tt\small yuanyang@ust.hk \quad hbliang@ust.hk$^{\dagger}$ \quad yichengwang@ust.hk$^{\dagger}$}
\end{tabular}
}

\maketitle

\section{Evaluation Protocol and Alignment Validation}

This supplementary material provides additional evaluation details and results for the main paper. All scene names, file paths, and site identifiers are omitted to preserve anonymity. We use the same terminology as the main paper and refer to the two evaluated sequences as the primary scene and the secondary scene.

All geometry evaluations are computed only on pixels supported by the rendered facade shell valid mask. This protocol evaluates visible facade-shell consistency on held-out views, rather than complete-scene reconstruction quality. Pixels corresponding to sky, dynamic objects, vegetation, occlusions, interiors, or regions outside the available exterior shell are excluded from the geometry metrics.

Rendered Gaussian normals and shell-rendered normals are compared in the camera coordinate frame after unit normalization. Normal error is reported as unsigned angular error, computed from the absolute dot product between the two unit normals. This evaluates local facade orientation rather than the normal-facing direction, which is the geometric property most relevant for our shell-supported reconstruction diagnostic. D-MAE is reported as a per-view median-scale-aligned depth diagnostic on shell-supported pixels, rather than as raw absolute metric-depth accuracy.

The exterior facade shell is aligned to the video reconstruction coordinate frame before shell-supervised training and then kept fixed. We validate the alignment by rendering the aligned shell into representative camera views and checking whether shell-supported regions overlap the visible facade structure. This validation is performed before optimization and is not tuned using final reconstruction results. The shell is therefore used as a validated structural prior for shell-supported regions, rather than as an absolute metric reference for the complete scene.

\begin{table}[t]
\centering
\small
\setlength{\tabcolsep}{6pt}
\caption{Ablation of shell-guided loss terms on the primary scene.
Geometry metrics are computed on shell-supported held-out test views.
D-MAE is per-view median-scale-aligned depth MAE in meters and is used
as a depth diagnostic. Normal is unsigned mean angular error in degrees;
threshold percentages are higher better.}
\label{tab:supp_ablation}
\begin{tabular}{lcccc}
\toprule
Variant & D-MAE & Normal & $<10^\circ$ & $<20^\circ$ \\
\midrule
Photo-only     & 0.236          & 55.21          & 2.30           & 7.71 \\
Depth only     & \textbf{0.113} & 55.42          & 1.84           & 6.74 \\
Normal only    & 0.183          & 17.74          & 47.64          & 72.33 \\
Depth + normal & 0.231          & \textbf{17.09} & \textbf{47.70} & \textbf{73.80} \\
Full objective & 0.273          & 17.42          & 47.36          & 73.12 \\
\bottomrule
\end{tabular}
\end{table}

\section{Primary-scene Ablation}

Table~\ref{tab:supp_ablation} reports the complete ablation of shell-guided loss terms on the primary scene. Depth-only supervision improves the scale-aligned depth diagnostic but does not correct facade orientation. In contrast, shell-rendered normal supervision is the dominant cue for reducing unsigned angular normal error. Combining depth and normal supervision gives the best image-space orientation metrics in this ablation, while adding the auxiliary geometry-consistency regularizer gives comparable but not better shell-supported image-space geometry. The main paper adopts the full objective as the default model because it retains the auxiliary geometry-consistency regularizer of the underlying reconstruction backbone while attaining image-space orientation comparable to the depth+normal variant. The shell-rendered depth and normal terms remain the dominant source of the orientation improvement.

\section{Spatial-region Holdout Stress Test}

Because our geometry metrics use the exterior facade shell as the visible-structure reference, we additionally conduct a spatial-region holdout stress test on the primary scene. We split shell-supported facade pixels into alternating vertical bands. The model denoted Ours-A receives shell supervision only on region A, while all metrics in Table~\ref{tab:spatial_holdout} are computed only on the disjoint region B, which is never used by shell-guided losses. The RGB reconstruction loss is still applied using the standard training views; only the shell-guided depth, normal, and geometry-consistency losses are spatially restricted to region A.

\begin{table}[t]
\centering
\small
\setlength{\tabcolsep}{5pt}
\caption{Spatial-region holdout stress test on the primary scene.
Ours-A receives shell-guided supervision only on region A, while all
metrics are computed only on the disjoint region B. All methods in this
table are evaluated on region B only; ``all regions'' denotes the
training supervision region, not the evaluation region. D-MAE is a
per-view median-scale-aligned depth diagnostic, and Normal is unsigned
mean angular error in degrees.}
\label{tab:spatial_holdout}
\begin{tabular}{lcccc}
\toprule
Method & D-MAE$\downarrow$ & Normal$\downarrow$ & $<10^\circ\uparrow$ & $<20^\circ\uparrow$ \\
\midrule
Photo-only         & 0.228 & 55.78 & 1.93  & 6.78 \\
MONO               & 0.209 & 54.82 & 6.86  & 13.59 \\
Ours (all regions) & 0.249 & 15.75 & 50.83 & 77.46 \\
Ours-A (A only)    & 0.170 & 49.18 & 5.82  & 14.86 \\
\bottomrule
\end{tabular}
\end{table}

The model trained with shell guidance on all regions remains strong on region B because it receives shell supervision there during training. In contrast, Ours-A improves depth error but provides only a limited normal-orientation gain over Photo-only and MONO on the never-supervised region B. This result indicates that shell-guided supervision is strongest on directly supervised shell-supported regions, while spatial extrapolation to disjoint unsupervised facade bands remains limited. Therefore, our evaluation should be interpreted as visible facade-shell consistency under available shell supervision, not as evidence of shell-only completion for unseen facade regions.

\section{Secondary-scene Shell Supervision}

\begin{figure}[t]
\centering
\includegraphics[width=0.52\linewidth]{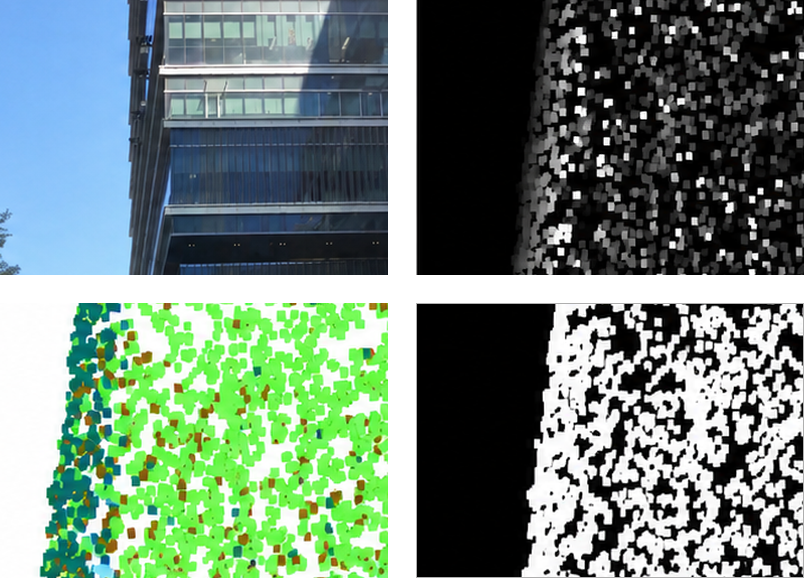}
\caption{Representative shell supervision on the secondary scene. The panels show, from top-left to bottom-right, the RGB frame, shell depth, camera-space shell normal, and valid mask.}
\label{fig:supp_secondary_supervision}
\end{figure}

Figure~\ref{fig:supp_secondary_supervision} shows a representative shell supervision example from the secondary scene. The visualization includes the RGB frame, shell depth, camera-space shell normal, and valid mask. White pixels in the valid mask denote shell-supported regions used for geometric supervision.

\section{Secondary-scene Quantitative Validation}

Table~\ref{tab:supp_secondary_geom} reports secondary-scene geometry metrics on shell-supported held-out test views. The secondary scene is used as a validation case rather than a large-scale generalization benchmark. Consistent with the primary scene, the strongest improvement appears in facade orientation: shell-guided supervision reduces unsigned normal error from 62.20 degrees to 23.44 degrees and increases the fraction of pixels below 20 degrees. The D-MAE diagnostic remains higher than the baselines, which is consistent with our scale-aligned diagnostic setting and the inverse-depth NCC training objective.

\begin{table}[t]
\centering
\small
\setlength{\tabcolsep}{6pt}
\caption{Secondary-scene geometry validation. Metrics are computed on
shell-supported held-out test views. D-MAE is per-view
median-scale-aligned depth MAE in meters and is used as a depth
diagnostic. Normal is unsigned mean angular error in degrees; threshold
percentages are higher better.}
\label{tab:supp_secondary_geom}
\begin{tabular}{lcccc}
\toprule
Method & D-MAE & Normal & $<10^\circ$ & $<20^\circ$ \\
\midrule
Photo-only & 1.637          & 62.20          & 1.18           & 4.10 \\
MONO       & \textbf{1.557} & 61.77          & 4.58           & 15.89 \\
Ours       & 1.889          & \textbf{23.44} & \textbf{43.47} & \textbf{63.43} \\
\bottomrule
\end{tabular}
\end{table}

\section{Secondary-scene Qualitative Geometry}

\begin{figure}[t]
\centering
\includegraphics[width=\linewidth]{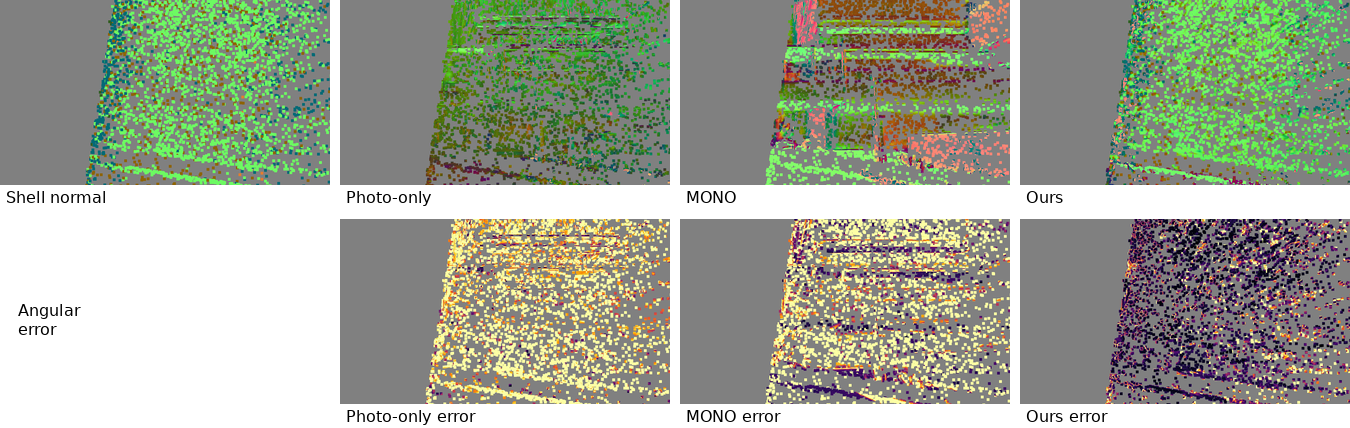}
\caption{Secondary-scene qualitative geometry comparison on a held-out
view. The top row compares camera-space normal maps, and the bottom row
shows unsigned angular normal-error maps on shell-supported regions.}
\label{fig:supp_secondary_qualitative}
\end{figure}

Figure~\ref{fig:supp_secondary_qualitative} provides a qualitative geometry comparison on a held-out secondary-scene view. To avoid street-level identifying content, the figure compares only rendered normal maps and angular-error maps on shell-supported regions. The shell-guided model produces more coherent facade orientation and lower unsigned angular-error regions than the photo-only and monocular-cue baselines.

